%% file: main.tex
\pdfoutput=1 
\documentclass{article}
\usepackage{siunitx}

\usepackage[utf8]{inputenc}
\usepackage[margin=0.75in]{geometry} %
\usepackage{authblk} %
\usepackage{lineno}
\usepackage{bbm}
\usepackage{newunicodechar}
\usepackage{mathtools}

\usepackage[backend=biber,style=nature,sorting=none]{biblatex} %
\let\cite=\supercite

\usepackage{algorithm,amsmath}

\usepackage{hyperref} 
\usepackage{cleveref}
\usepackage{xcolor}
\hypersetup{
  colorlinks   = true, %
  urlcolor     = blue, %
  linkcolor    = blue, %
  citecolor   = blue %
}

\usepackage[noend]{algpseudocode}
\usepackage{graphicx}
\usepackage{float}
\usepackage[labelformat=simple]{subcaption} 
\usepackage[hypcap=true]{caption}
\usepackage[T1]{fontenc}
\usepackage{lmodern}
\usepackage[labelfont=bf]{caption}

\title{Improving debris flow evacuation alerts in Taiwan using machine learning}

\author[1,*]{Yi-Lin Tsai}
\author[2]{Jeremy Irvin}
\author[2]{Suhas Chundi}
\author[2]{Andrew Y. Ng} 
\author[3,4,5]{Christopher B. Field}
\author[1,3,6]{Peter K. Kitanidis}

\affil[1]{Department of Civil and Environmental Engineering, Stanford University, Stanford, CA, USA}
\affil[2]{Department of Computer Science, Stanford University, Stanford, CA, USA}
\affil[3]{Woods Institute for the Environment, Stanford University, Stanford, CA, USA}
\affil[4]{Interdisciplinary Environmental Studies Program, Stanford University, Stanford, CA, USA}
\affil[5]{Department of Earth System Science, Stanford University, Stanford, CA, USA}
\affil[6]{Institute for Computational and Mathematical Engineering, Stanford University, Stanford, CA, USA}
\affil[*]{yilin2@stanford.edu}

\begin{document}

\date{\today}
\vspace{-0.5in}

\maketitle

\begin{abstract}

Taiwan has the highest susceptibility to and fatalities from debris flows worldwide. The existing debris flow warning system in Taiwan, which uses a time-weighted measure of rainfall, leads to alerts when the measure exceeds a predefined threshold. However, this system generates many false alarms and misses a substantial fraction of the actual debris flows. Towards improving this system, we implemented five machine learning models that input historical rainfall data and predict whether a debris flow will occur within a selected time. We found that a random forest model performed the best among the five models and outperformed the existing system in Taiwan. Furthermore, we identified the rainfall trajectories strongly related to debris flow occurrences and explored trade-offs between the risks of missing debris flows versus frequent false alerts. These results suggest the potential for machine learning models trained on hourly rainfall data alone to save lives while reducing false alerts.

\end{abstract}


\input{intro}

\input{results}

\input{discussion}

\input{methods}

\section*{Data availability}
The datasets generated during and/or analyzed during the current study are available from the corresponding author on reasonable request.

\printbibliography
\section*{Acknowledgements}

\textit{Technical Supports:} We thank Dr. Mary McDevitt of the Technical Communication Program in the School of Engineering at Stanford University for rigorously reviewing and revising our manuscript, and providing constructive feedback and valuable instructions in detail. We also thank João Pedro Araújo for his advice and exploratory assistance in the early stage of this study.

\textit{Funding:} This research was supported by the Department of Civil and Environmental Engineering and Department of Computer Science at Stanford University, AI for Climate Change Bootcamp at the Stanford Machine Learning Group, Microsoft AI for Earth program, and Stanford Artificial Intelligence Lab.

\section*{Author contributions}
Y.L.T., J.I., P.K.K., and C.B.F. designed the research; all authors developed and refined the methodologies of analyses; Y.L.T., J.I., and S.C. performed the analyses; Y.L.T., J.I., A.Y.N., C.B.F., and P.K.K. interpreted results. Y.L.T. wrote the first draft of the manuscript. Y.L.T., J.I., A.Y.N., C.B.F., and P.K.K. revised the manuscript. All authors reviewed the final version of the manuscript.

\section*{Ethics declarations}
\subsection*{Competing interests}
The authors declare no competing interests.

\section*{Disclaimer}
This disclaimer informs readers that this study should only serve as their own references rather than official guidelines for any jurisdictions. The scenarios of simulations performed in this study are only examples for the purpose of academic research. Any actions, including but not limited to decisions, policies, and studies, taken based on any part of this study is the sole liability of readers, not authors in this study.

\end{document}

%% file: intro.tex
\section*{Introduction}

Disasters due to extreme precipitation and associated compound hazards, such as floods, landslides, and debris flows, have been increasing around the world \cite{arsix2021, moftakhari2019increasing, wahl2015increasing}. Global climate change is projected to increase the intensity and frequency of heavy rainfalls, exacerbating these events \cite{arsix2021, field2012managing}. One of the most frequent compound disasters is rainfall-triggered debris flows, which are large amounts of water mixed with soil, rocks, and wood rushing down the slope of mountains, often causing widespread damage and fatalities. Effective management of the risk of complex geological hazard cascades should be a top priority \cite{cascades2021}, including effective ways to anticipate the occurrences of debris flows to enable timely evacuations.  
   
Taiwan has the highest level of landslide susceptibility \cite{stanley2017heuristic} and fatalities \cite{kirschbaum2015spatial} in the world. The frequency and magnitude of debris flows have increased in Taiwan due to extreme rainfall brought by seasonal typhoons (also known as $hurricanes$ or $cyclones$) and loose colluvium that results from frequent earthquakes \cite{chen2016landslide, jan2005debris}. Compared to topography, soil conditions, and geology, rainfall is the most influential contributor to the occurrences of debris flows \cite{jan2018txt, chae2016feasibility, chen2015rainfall, chen2013characteristics, chen2012recent}. Therefore, the Soil and Water Conservation Bureau (SWCB) of Taiwan uses a rainfall-based debris flow warning model \cite{jan2018txt} to alert local communities to evacuate when a critical threshold value of the effective accumulated rainfall (calculated as in equation (\ref{eq:effective_antecedent_rainfall})) is exceeded (see Methods). 

Predicting the occurrence of debris flows is challenging due to the complex and stochastic relationship between meteorological and topographical variables and debris flow occurrences \cite{chang2007application}. Based on our analysis of the case reports of major debris-flow-related disasters in Taiwan \cite{major}, the existing model issued an evacuation alert prior to the debris flow in only $20\%$ of debris flows since 2010, using the predetermined critical value of the effective accumulated rainfall at each debris-flow-prone watershed. $55\%$ of the debris flows occurred before the predetermined critical values were exceeded, and $25\%$ of the debris flows occurred in a watershed where a critical value had not been established. 

Some studies have used radar or satellite rainfall products to predict precipitation and debris flows \cite{nikolopoulos2017satellite, marra2016space, marra2014radar}. While promising, these approaches tend to suffer from high uncertainty, limited data availability, and high cost \cite{nikolopoulos2017satellite, marra2014radar}. Data-driven machine-learning models are increasingly employed in improving early warning models, weather and natural hazard forecasts, and disaster evacuation management. Examples include weather forecasting \cite{knusel2019applying, lin2009effective}, landslide displacement prediction \cite{yang2019time}, spatial mapping of debris flow susceptibility \cite{ng2021spatiotemporal, chen2020spatial, zhang2019debris, kirschbaum2018satellite}, predicting scales of landslides \cite{huang2018method} and monthly rainfall for early warning of landslide occurrence \cite{srivastava2020monthly}, differentiating between ground vibrations generated by debris flows and other seismic signals \cite{chmiel2021machine}, and enhancing disaster response and emergency evacuation planning \cite{cegan2022importance, tsai2021routing, deparday2019machine, tsai2019deep}. However, to the best of our knowledge, none of the existing studies predict the occurrences of debris flows within a selected time using machine learning algorithms trained on historical hourly rainfall data alone.
      
This study used Taiwan as a case study to investigate whether data-driven machine learning methods can enhance the efficacy of the government-endorsed and widely-used rainfall-based debris flow warning model. We used precipitation data from the Central Weather Bureau (CWB) \cite{cwb} and the Data Bank for Atmospheric and Hydrologic Research \cite{databank}, plus case reports \cite{major} of major debris-flow-related disasters. These reports include data on debris flows, landslides, erosion, floods, and subsidence from the SWCB of Taiwan. We included all 58 weather stations in Taiwan where at least one debris flow occurred between 2015 and 2020. From these stations, we collected rainfall data around the dates of all 104 debris flow events (positive datasets). We also gathered 533 additional rainfall data with at least one period of rain over 4 mm h-1 for at least two consecutive hours, but without occurrence of debris flow (negative datasets). The consolidated precipitation dataset includes 637 datasets, each with 210 hours of rainfall data on average. These datasets correspond to 136,320 pairs of hourly rainfall values and labels indicating whether or not there was a debris flow event within the subsequent 12 hours, which we call positive and negative labels respectively.

We used this dataset to train five machine learning models (see Methods): Logistic Regression (LR) \cite{pampel2020logistic, kleinbaum2002logistic}, Multi-layer Perceptron (MLP) \cite{taud2018multilayer, bishop1995neural}, Random Forest (RF) \cite{biau2016random, cutler2012random, biau2012analysis, scikit-learn, hastie2009random, breiman2001random}, Extreme Gradient Boosting (XGBoost) \cite{chen2016xgboost, chen2015xgboost}, and Light Gradient Boosting Machine (LightGBM) \cite{ke2017lightgbm}, to predict the occurrence of a debris flow within the upcoming 12 hours. We used $85\%$ of the data as a training set to allow each machine learning model to learn parameters and $15\%$ of the data as a test set to evaluate the performance of the best machine learning model against the ETM and the HM on previously unseen data. We compared the performance of machine learning models using six different combinations of input data in the $85\%$ training set. All six combinations included only hourly rainfall values. 

We compared the performance of the best machine learning model against the current warning model used in Taiwan, referred to here as the Existing Taiwan Model (ETM) adopted by the SWCB of Taiwan, and a Homogeneous Model (HM) developed in this study. The ETM includes an effective accumulated rainfall (EAR) threshold value for each debris-flow-prone watershed \cite{jan2018txt} (see the detailed definition of EAR in Methods). The HM differs from the ETM in that a uniform threshold value of EAR is used. 

\begin{figure}[H]

\includegraphics[width=0.8\textwidth]{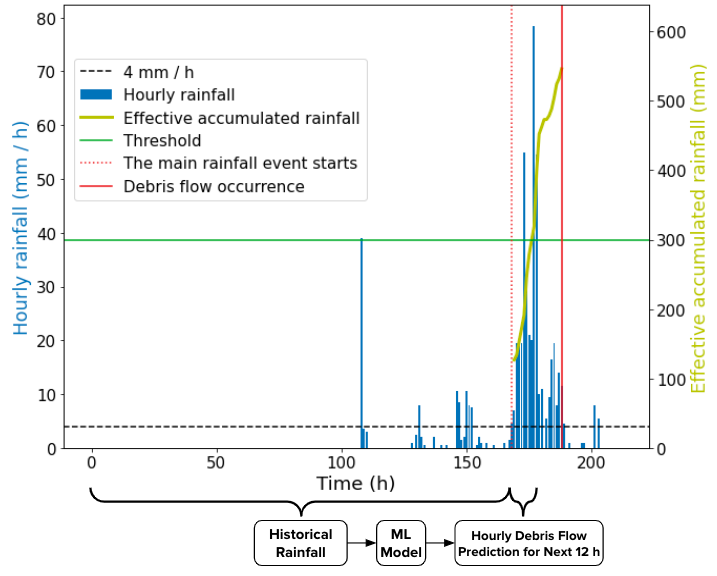}\centering
\caption{\textbf{The mechanism of evacuation alerts in the current rainfall-based debris-flow warning model and the machine learning (ML) models.} The time series associated with a rainfall event with debris flow occurrence at the Shangguguan Station on May 19, 2019 illustrates how the current rainfall-based debris-flow warning model works. The blue bars represents the hourly rainfall in mm/h. The yellow curve indicates the effective accumulated rainfall, measured on the right y-axis. The black dashed line indicates the threshold of 4 mm h-1 that marks the starting point of a rainfall event. The dashed red vertical line indicates the first hour of the main rainfall event. The EAR (the yellow curve calculated from equation (\ref{eq:effective_antecedent_rainfall})) increases over time starting from the first hour of the main rainfall event. When it exceeds 300 mm (horizontal green line), which is the EAR threshold value determined by experts at this station, the debris flow evacuation warning is triggered. The solid red vertical line shows when the debris flow actually occurred. In this case, the ETM successfully issued the evacuation warning 10 h ahead of the debris flow. The text boxes below the x-axis show that the machine learning models in this study use hourly rainfall values as inputs to predict on an hourly basis whether a debris flow will occur within the subsequent 12 h.}\label{fig:input_output_example}
\centering
\end{figure}

An actual rainfall event with a debris flow occurrence (Fig. \ref{fig:input_output_example}) provides an example of how the ETM and the machine learning models work in real-world evacuation alerts. The machine learning models use the same time-stamped hourly rainfall information used in the ETM. Instead of aggregating the hourly rainfall data into the EAR which is used as the input in the ETM (see Methods), the machine learning models simply use hourly rainfall values as inputs to predict on an hourly basis whether a debris flow will occur within the subsequent 12 h. In Supplementary Information, we compare the results of using hourly rainfall values as the only inputs with those that use composites of the hourly rainfall that include EAR and daily rainfall values. If the model predicts that a debris flow will occur in the subsequent 12 h, the warning is triggered ($positive$ prediction); otherwise, no warning is issued ($negative$ prediction).

The machine learning models and the Taiwan models output a continuous sequence of scores that need to be converted to binary values to determine whether to issue an alert. The output of the machine learning models is the predicted probability $P_{predict}$ between $0$ and $1$ of a debris flow within the upcoming 12 h. To convert the predicted probability to either \textit{positive} (a debris flow will occur within the upcoming 12 h) or \textit{negative} (a debris flow will not occur within the upcoming 12 h), we used threshold $P_{threshold}$ for all debris-flow-prone watersheds. If $P_{predict} \ge P_{threshold}$, the machine learning models output \textit{positive}. Otherwise, the machine learning models output \textit{negative}. The ETM and the HM utilize an effective accumulated rainfall (EAR) threshold value for issuing a warning. Once the EAR threshold is exceeded, the ETM and the HM generate \textit{positive}. Otherwise, they generate \textit{negative}. The higher the values of the $P_{threshold}$ and EAR threshold, the rarer evacuation alerts are issued.

To measure model performance, we used standard evaluation metrics including precision $\frac{TP}{TP+FP}$, recall $\frac{TP}{TP+FN}$, and specificity $\frac{TN}{TN+FP}$ to measure the performance in making binary predictions (Fig. \ref{fig:confusion_matrix}). Precision is the proportion of predicted alerts for which a debris flow actually occurred, recall quantifies the proportion of debris flow events for which an alert was issued, and specificity quantifies the proportion of negative events for which no alert was issued. Each value of $P_{threshold}$ and EAR threshold leads to different values of precision, recall, and specificity. To investigate the implications of threshold value, we tested a range of thresholds in the ETM, the HM, and the machine learning models. By varying $P_{threshold}$ from 0 to 1 for the machine learning models, we can construct a curve. The Receiver Operating Characteristic (ROC) curve plots 1 - specificity against recall and the Precision-Recall (PR) curve plots precision against recall. For the ETM and HM model, varying the EAR threshold from 0 to the maximum EAR value, either in proportion to the actual threshold (for ETM) or as a uniform value across stations (for HM), allows us to construct conceptually similar curves.

To summarize the performance of the model across all thresholds, we use the area under the PR curve (referred to as AUPRC) and the area under the ROC curve (referred to as AUROC). The values of AUPRC and AUROC range from 0 to 1, where larger AUPRC and AUROC values indicate better model performance. The better the model performance, the closer the ROC curve is to the upper left corner and the closer the PR curve is to the upper right corner. A ``no skill'' model predicts randomly so its AUROC is 0.5 and its AUPRC depends on the proportion of positive examples in all examples. When positive examples are rare compared to negative examples, the AUPRC of the ``no skill'' model is close to 0. We used AUPRC to evaluate the machine learning models and to compare the best one against the Taiwan models as AUPRC is sensitive to the low prevalence of positive events compared to negative events. However, AUROC does not depend on the ratio of positive events to negative events. Furthermore, AUROC is inflated by a large number of negative events due to a high number of true negatives, whereas achieving high AUPRC requires a high number of true positives as well as a low number of false positives and false negatives.  Finally, we evaluated the variability around the estimated AUPRC due to the finite size of the sample. For the ETM, the HM, and the best machine learning model on the test set, we used circular block bootstrapping \cite{circular_block_bootstrap, politis1991circular} on the test set with 10000 bootstrap replicates. To account for correlation in the hourly rainfall data, the circular block bootstrapping sampled by 6-h blocks rather than by hours.

%% file: results.tex
\section*{Results}

\subsection*{Performance comparison of machine learning models}
To compare the performance of the five machine learning models, we performed a 10-fold cross-validation on the $85\%$ training set. This means that $90\%$ of the data was used to train the machine learning model and the remaining $10\%$ to evaluate its prediction performance, repeated $10$ times each on different random samples of the training set. Then, we used the mean of the areas under the PR curves (AUPRC) from the 10-fold cross-validation as the estimate of model performance. We found that the random forest model (RF) outperformed the other four machine learning models for rainfall inputs ranging from the most recent six hours to the most recent 168 hours on the training set (Table \ref{table:cv_scores_ml_models}). Across the six input data options in Table \ref{table:cv_scores_ml_models}, cross validation scores generally increased with the length of the rainfall record and then saturated. RF, which had the highest cross validation scores, overall, saturated with the most recent 96 hourly rainfall values (H=96 in Table \ref{table:cv_scores_ml_models}).

\begin{table}[H]

\caption{\textbf{Performance comparison between five machine learning models using the cross-validation scores (AUPRC).}}\label{table:cv_scores_ml_models}
\centering
\includegraphics[width=0.66\textwidth]{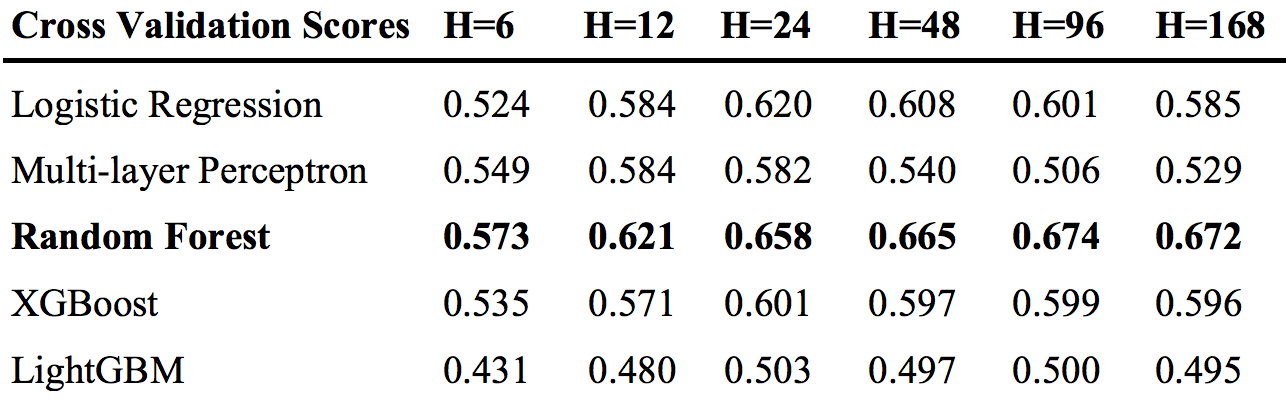}\centering
\caption*{The hourly rainfall values in the most recent 6, 12, 24, 48, 96, or 168 hours were used as inputs in the machine learning models. $H$ denotes the number of hours.}
\end{table}

\subsection*{Comparison between the RF and the Taiwan models}

\subsubsection*{Performance evaluation}
\begin{figure}[H]\captionsetup{singlelinecheck = false, justification=justified}

	\centering
	\begin{subfigure}{0.49\textwidth} %
	    \caption{Receiver operating characteristic curves (ROC curves)} %
		\includegraphics[width=\textwidth]{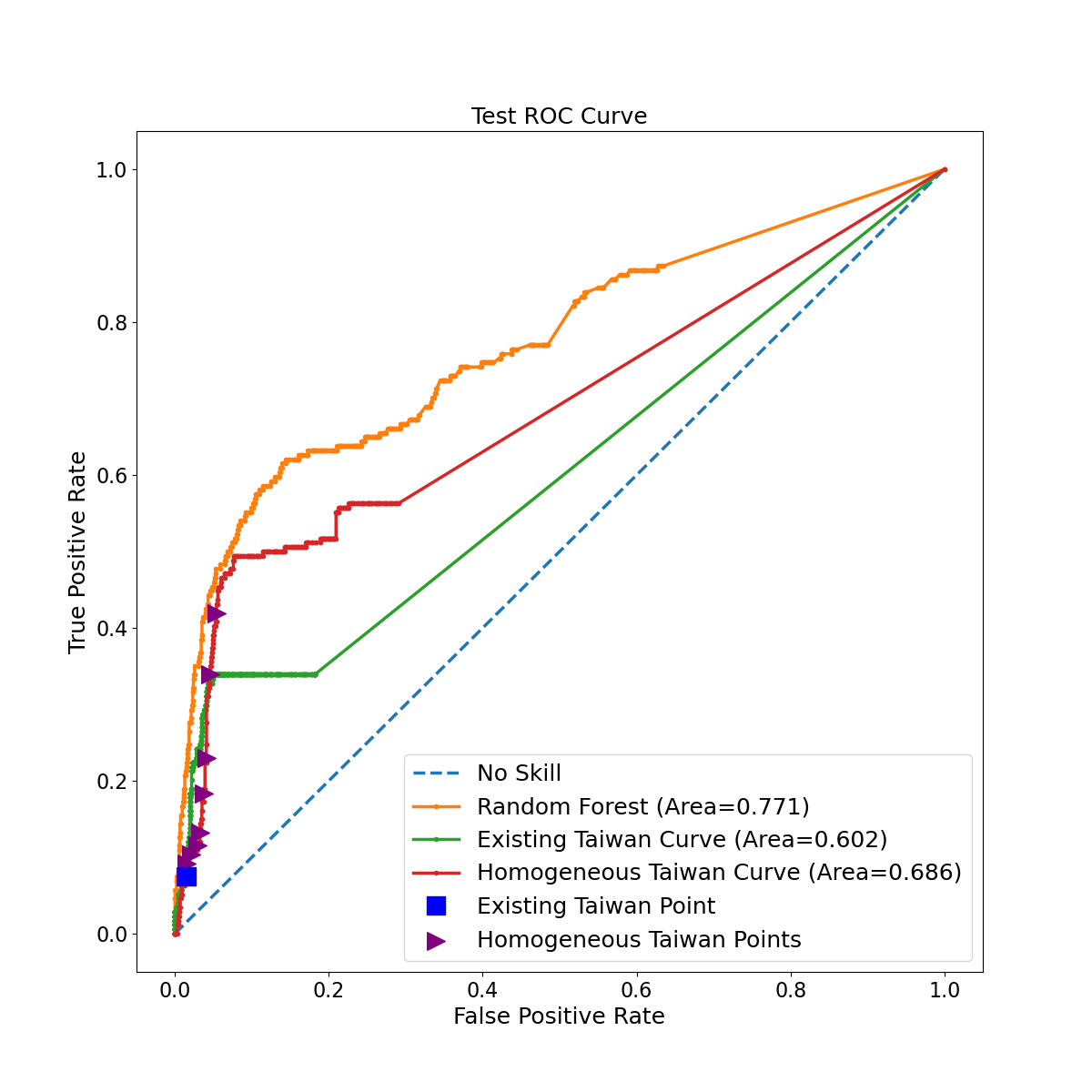}
	\end{subfigure}
	\hspace{0cm} %
	\begin{subfigure}{0.497\textwidth} %
	    \caption{Precision-Recall curves (PR curves)} %
		\includegraphics[width=\textwidth]{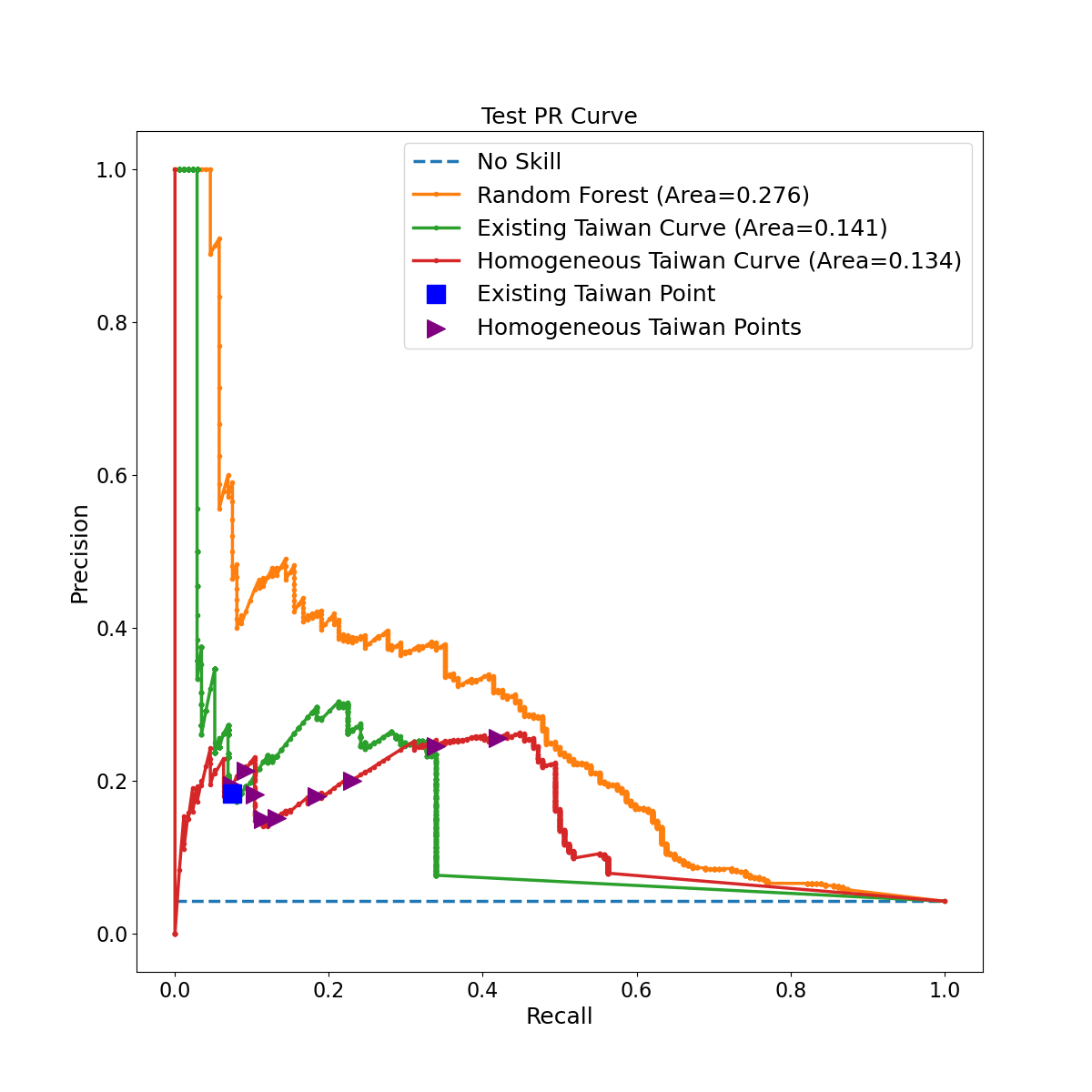}
	\end{subfigure}
	\caption{\textbf{Comparison between the performance of the ETM, the HM, and the RF on the test set.} Individual points indicate different thresholds of EAR for the ETM and the HM and thresholds of probabilistic values for the RF (see Methods). The blue squares indicate the performance of the ETM using the expert-determined thresholds for EAR. We changed these values in steps of $0.1\%$ to establish the green ROC and PR curves. The purple triangles signify nine uniformly spaced EAR thresholds, ranging from 200 to 600 mm, with an interval of 50 mm. This encompasses the range of expert-determined thresholds in the ETM. To generate the curves for the HM, we used the same EAR threshold at all sites, gradually increasing it from zero until the maximum EAR value. The blue dashed lines show the performance of a model making random predictions (``No Skill'' model).}\label{fig:roc_pr}  
\end{figure}

We compared the performance of the RF, using hourly rainfall values in the most recent 96 hours as inputs, against the ETM and the HM on the $15\%$ test set (Fig. \ref{fig:roc_pr}), the portion of the data that was not seen by the machine learning models during the training process. The RF outperformed the ETM and the HM across the entire ROC and PR curves (Fig. \ref{fig:roc_pr}). In the ROC curves, when the false positive rate was below 0.05, the true positive rate of the RF was the highest, followed by the ETM and then the HM (Fig. \ref{fig:roc_pr}a). When the false positive rate was greater than 0.05, the HM outperformed the ETM and was comparable to the RF for false positive rates between 0.05 and 0.08. Although the ROC curves of the RF, the ETM, and the HM looked very close to each other for any false positive rate below 0.08, under the same level of false positive rate, the RF had the highest true positive rate, which can be two to three times as high as the true positive rates of the ETM and the HM.

In the PR curves, the RF substantially outperformed the ETM and the HM across the whole range of threshold values (Fig. \ref{fig:roc_pr}b). The AUPRC values were 0.276 ($95\%$ CI 0.160, 0.406) for the RF, 0.141 ($95\%$ CI 0.068, 0.240) for the ETM, and 0.134 ($95\%$ CI 0.054, 0.187) for the HM. Although the $95\%$ confidence intervals of these three models overlap, the RF has a substantially higher probability of providing accurate debris flow evacuation alerts on the $15\%$ test set than the ETM or the HM.

\subsubsection*{Operating points}

In practice, the SWCB selects a single threshold to decide whether to issue an alert. Selecting this threshold is difficult as there is a fundamental trade-off between false alerts and failures to issue early evacuation warnings. Furthermore, each country has a different tolerance for erroneous and missed alerts. In Japan, for example, warnings are issued quite often to avoid missing potential debris flows, but this leads to frequent false alerts \cite{japan}. Taiwan, on the other hand, tends to issue debris flow evacuation warnings conservatively, only when the probability of a debris flow is high, but this leads to more failures to issue warnings prior to actual debris flows. Our study confirms that the ETM-based debris flow warning model is conservative, since its precision is higher than its recall in the PR curves when the ETM uses the expert-determined thresholds for EAR (the blue square in Fig. \ref{fig:roc_pr}b).

To investigate the number of false positive alerts at specified levels of captured debris flow events, we evaluate the performance of the models at various values of recall (Table \ref{table:precision_recall_corresponding}). Specifically, for recall values of 0.1 to 0.9, we measure the precision (the fraction of the alerts that are actual debris flows) and the specificity (the fraction of the events without debris flow occurrences that are correctly predicted). We also measure these performance metrics at the current ETM thresholds determined by hydrologists and disaster managers (the \textit{current} column in Table \ref{table:precision_recall_corresponding}a). The RF outperformed the ETM and the HM in both precision and specificity. At the recall of the ETM (0.075), the RF had the highest precision and specificity, followed by the HM, and the ETM was the lowest (see the first column in Table \ref{table:precision_recall_corresponding}a).

\begin{table}[H]\captionsetup{singlelinecheck = false, justification=justified}

\caption{\textbf{Trade-offs for the Existing Taiwan model, the Homogeneous model, and the RF model, showing the corresponding precision and specificity across a range of values of recall and the corresponding recall and specificity across a range of values of precision.}}\label{table:precision_recall_corresponding}   
	\centering
	\begin{subtable}{0.66\textwidth} %
	    \caption{Corresponding precision and specificity based on recall} %
		\includegraphics[width=\textwidth]{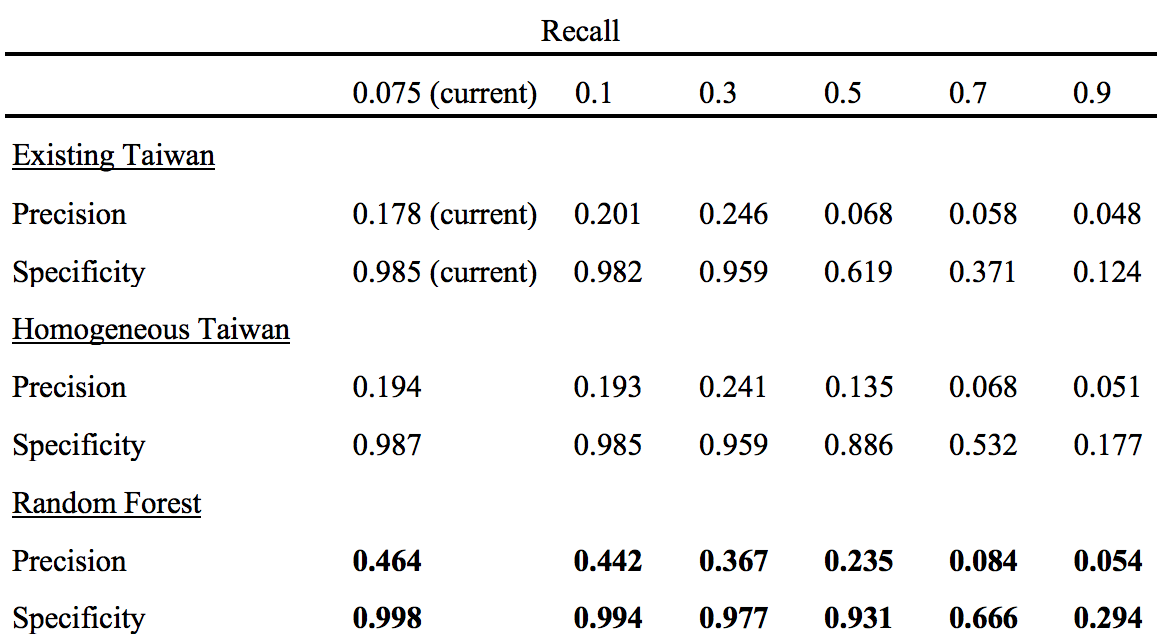}
		\vspace*{0.5mm}
	\end{subtable}
	
	\begin{subtable}{0.66\textwidth} %
	    \caption{Corresponding recall and specificity based on precision} %
		\includegraphics[width=\textwidth]{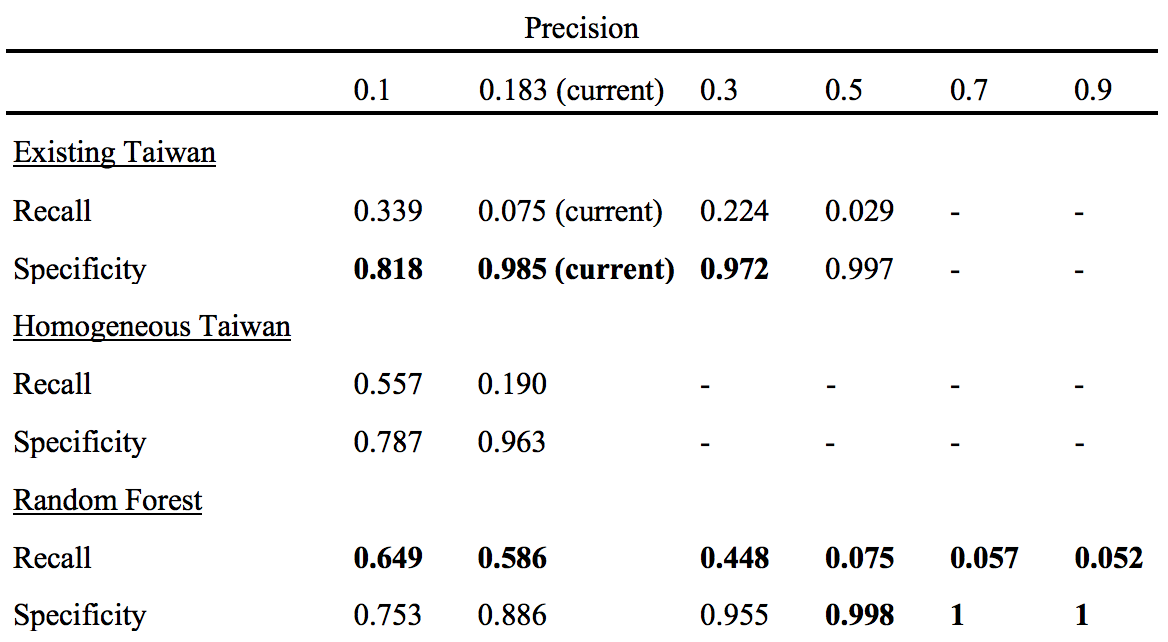}
	\end{subtable}
\caption*{These two tables can serve as a guide for selecting to determine operating points.}
\end{table}

To investigate the fraction of captured events at specified levels of false positive alerts, we evaluate recall (the fraction of events with debris flows that are correctly predicted) and specificity (the fraction of the events without debris flows that are correctly predicted) at various values of precision (Table \ref{table:precision_recall_corresponding}). We also report these metrics at the current ETM thresholds determined by the SWCB of Taiwan (the \textit{current} column in Table \ref{table:precision_recall_corresponding}b). The RF captured about 8 and 3 times as many debris flow occurrences as the ETM and the HM, respectively, when the precision value was 0.183. Moreover, the RF had a wider range of operating points, with thresholds that enabled achieving precision values up to 0.9 (Table \ref{table:precision_recall_corresponding}b). If there were no feasible thresholds under certain precision levels (e.g., HM for precision greater than 0.25) then the model did not have corresponding points along the PR curve (Fig. \ref{fig:roc_pr}b). The ETM generated the fewest false alerts (i.e., highest specificity) when precision was below 0.3, while the RF avoided the most false alerts when precision was greater than 0.5.

\subsection*{Interpretation of the RF model}
\subsubsection*{Important data inputs}

\begin{figure}[H]

\includegraphics[width=0.7\textwidth]{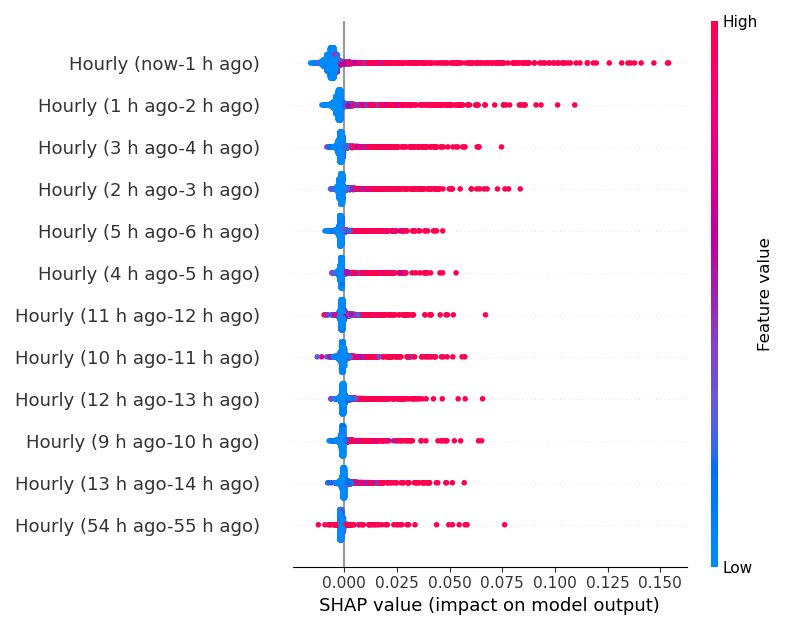}\centering
\caption{\textbf{SHAP values for the 12 most important data inputs when a trained RF made predictions for the $15\%$ test set.} Every point represents the SHAP value for one example \cite{lundberg2020local2global, lundberg2017unified}. Overlapping scattered points are moved in the direction of y-axis to show the distribution of the SHAP values. Data inputs are sorted in the order of importance across all examples. The vertical ``feature value'' color bar indicates the values of data inputs (i.e., rainfall values). The larger the positive SHAP value of a data input, the more that data input contributes to a positive prediction (debris flow within 12 h).}\label{fig:top12_shap_rf}
\centering
\end{figure}

To assess the contribution of each input to a debris flow prediction, we used a SHAP (SHapley Additive exPlanations) plot \cite{lundberg2020local2global, lundberg2017unified}. For each prediction in the test set, SHAP produces values that indicate the impact of each input on the prediction, where a positive value indicates the input contributed to a $positive$ prediction (a debris flow would occur within 12 h), and a negative value indicates the input contributed to a $negative$ prediction (no debris flow would occur within 12 h). For the RF with the most recent 96 hourly rainfall values as inputs (Fig. \ref{fig:top12_shap_rf}), the most important input was the hourly rainfall from the most recent hour (now--1 h ago).

In general, the most recent hourly rainfall values had high SHAP values (Fig. \ref{fig:top12_shap_rf}). Although the exact ranking varied across the 55 experiments, rainfall in the most recent hour always had a high SHAP value, suggesting that the most recent rainfall is the strongest determinant of debris flow. However, the trend in the SHAP plot does not necessarily imply causality \cite{molnar2022}. Instead, the SHAP plot describes the correlations between data inputs (i.e., features) and outputs (i.e., predictions) in a machine learning model.

\subsection*{Trade-offs among multiple objectives}
\subsubsection*{Predictions using different training weights}
\begin{figure}[H]\captionsetup{singlelinecheck = false, justification=justified}

	\centering
	\begin{subfigure}{0.48\textwidth} %
	    \caption{Trade-offs between FNR and FPR} %
		\includegraphics[width=\textwidth]{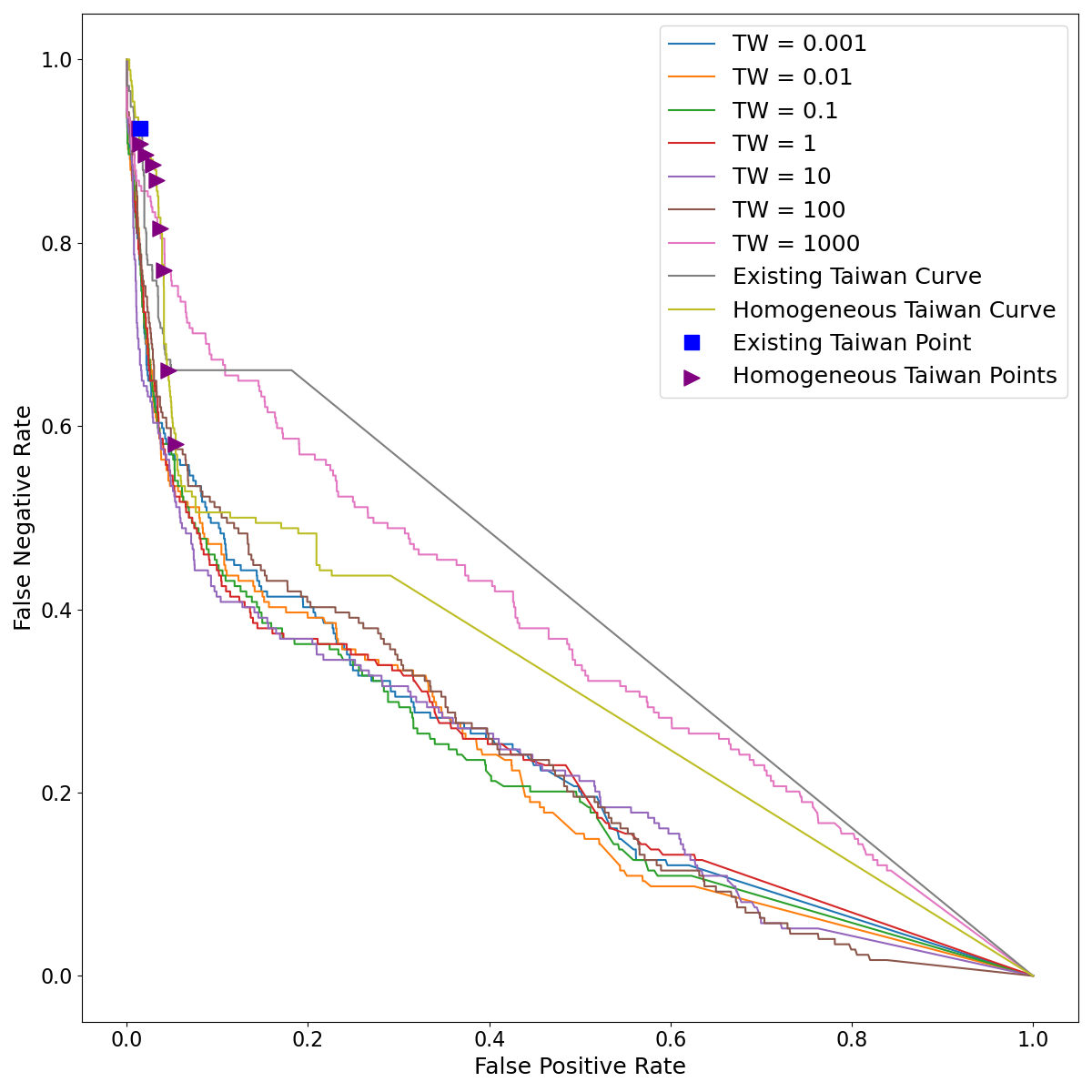}
	\end{subfigure}
	\hspace{0.3cm} %
	\begin{subfigure}{0.48\textwidth} %
	    \caption{Trade-offs between FOR and FDR} %
		\includegraphics[width=\textwidth]{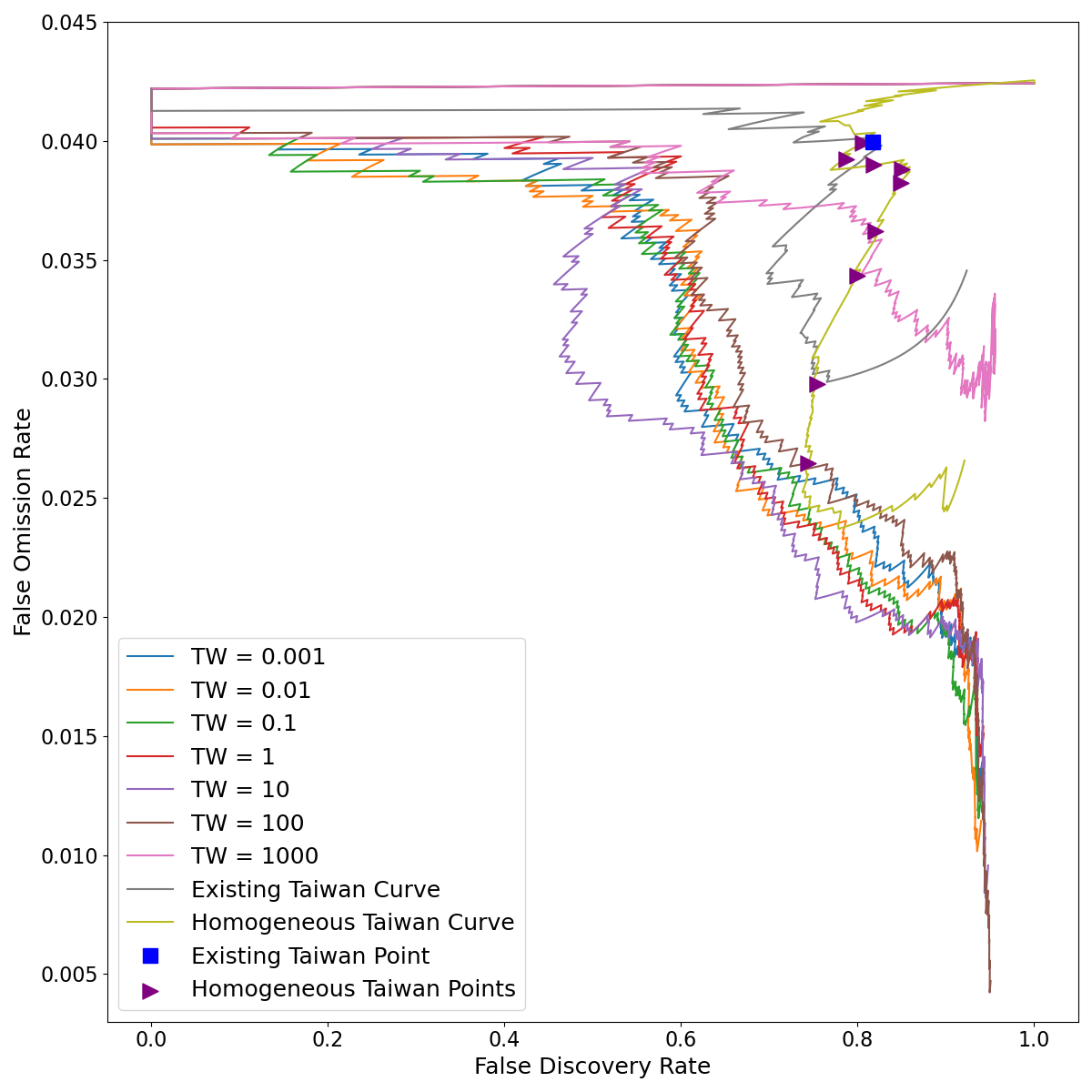}
	\end{subfigure}
	\caption{\textbf{Adjusting the training weights (TW) of the RF to value avoiding false alerts or missing warnings when a debris flow is imminent, showing consequences for trade-offs between false negative rate and false positive rate (Fig. \ref{fig:multi_objectives}a) and false discovery rate and false omission rate (Fig. \ref{fig:multi_objectives}b) using the $15\%$ test set.} For the RF, the ETM, and the HM, curves were generated by moving the thresholds for issuing evacuation warnings, as in Fig. \ref{fig:roc_pr}. The RF with a training weight = 1 was used for Figs. \ref{fig:roc_pr} and \ref{fig:top12_shap_rf}.}\label{fig:multi_objectives}  %
\end{figure}

To explore options for addressing alternative priorities with the RF, we trained the model with a range of weights on the prediction of a debris flow relative to no debris flow. A high weight trains the model to emphasize not missing debris flows. A low weight emphasizes avoiding unnecessary evacuations. Independent of the training weight, over a range of weights from $10^{-3}$ to $10^3$, the false positive rate, false negative rate, false omission rate, and false discovery rate (see Methods) of the RF were mostly lower than those of the ETM and the HM (Fig. \ref{fig:multi_objectives}).

The patterns in Fig. \ref{fig:multi_objectives} jumped around, and intermediate values led to the extremes in the performance metrics because adding weights can lead to training instabilities and unexpected behavior in the trade-offs, which may be caused by the relatively small dataset sizes. The patterns in Fig. \ref{fig:multi_objectives} were not useful for evacuation planning because there was no clear relationship between different training weights and the trade-offs between those undesired outcomes. With sufficient datasets, the curves in Fig. \ref{fig:multi_objectives} would become well behaved, and a disaster manager can use the trade-offs between those conflicting goals as a decision-making tool to choose a training weight under a preferred level of target objective to prepare for future debris flow evacuation alerts. For example, if a disaster manager wants to set up 0.03 as the false omission rate, choosing 10 as the training weight can be an optimal solution because the false discovery rate is the lowest, compared to other training weights.

\subsubsection*{Predictions using different thresholds}
\begin{figure}[H]

\includegraphics[width=0.7\textwidth]{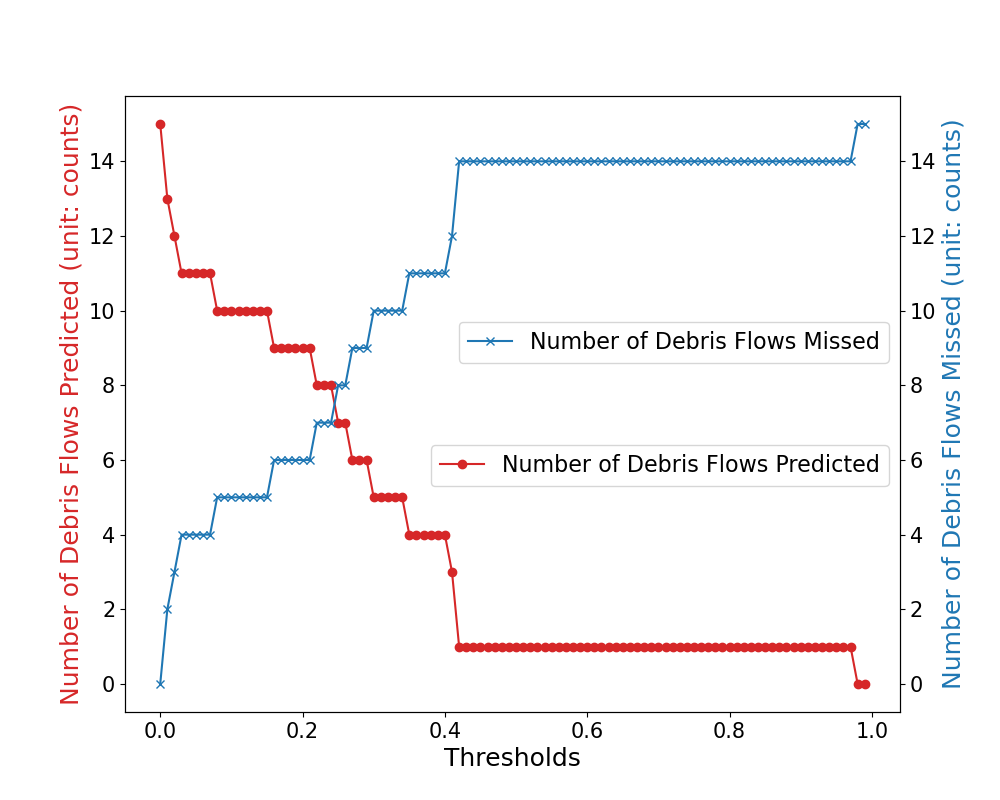}\centering
\caption{\textbf{The number of debris flows predicted or missed by the RF model using different thresholds.} Every point shows the number of debris flow events the RF predicted (left y-axis) or missed (right y-axis) within the 12-h lead time window using different probabilistic thresholds. The x-axis represents different probabilistic thresholds, ranging from 0 to 1, with an interval of 0.01. There are 15 debris flow events (positive datasets) in the $15\%$ test set.}\label{fig:Test_threshold_predict_analysis}
\centering
\end{figure}

To analyze the number of debris flows can be predicted within the 12-h time window before a debris flow occurred, the probabilistic thresholds, increasing from 0 to 1 with an increment of 0.01, were applied when we trained the RF, using hourly rainfall values in the most recent 96 hours as inputs, on the $15\%$ test set. As expected, the lower the threshold, the more debris flows were predicted by the RF at the expense of lower precision (Fig. \ref{fig:Test_threshold_predict_analysis} and the high recall, low precision in Fig. \ref{fig:roc_pr}b). When the ETM used the expert-determined
thresholds for EAR (the blue square in Fig. \ref{fig:roc_pr}b), it predicted two of 15 debris flow events in the $15\%$ test set. In other words, the performance of the ETM was close to the RF using 0.4 as the threshold. Disaster managers can use Fig. \ref{fig:Test_threshold_predict_analysis} as an analysis tool to determine a preferred threshold for future debris flow evacuations.

\subsection*{Impacts of various input data on model performance}
To explore the impacts of various input data on the performance of the RF, we tried 55 input data combinations of EAR, hourly precipitation, and daily precipitation as an ablation study (see Methods and Supplementary Information), including the six combinations in Table \ref{table:cv_scores_ml_models}. Daily precipitation and EAR were calculated from hourly precipitation values. In other words, we did not add new information but instead augmented the raw hourly rainfall data with the same information collected into daily packages (daily precipitation) or into a time-weighted package (EAR).

Regardless of whether the EAR was used as input data, using hourly rainfall values as the input data provided simpler input data combinations but similarly competitive model performance compared to other kinds of input data combinations (Supplementary Information). When EAR was used, it was one of the most important inputs (e.g., Fig. S31 in Supplementary Information), reflecting its utility in integrating and weighting rainfall history for a system like the ETM. However, when EAR was the only data point used to train the RF, the AUPRC was 0.380, much smaller than the average (0.642) of the other 27 experiments that included daily, hourly, or both daily and hourly rainfall values in addition to EAR. In contrast, adding the EAR in addition to daily, hourly, or both daily and hourly rainfall values as inputs only marginally improved the model performance by 2.66\% on average (Supplementary Information), demonstrating that the machine learning models can effectively recreate the predictive value that comes from EAR.

%% file: discussion.tex
\section*{Discussion}

Timely debris flow evacuation alerts are an economic and life-safety priority but a major scientific challenge. Effects of climate change on the frequency and intensity of extreme precipitation will likely increase both the importance and the challenges of issuing evacuation alerts that are accurate and timely. Existing systems have certainly saved lives, but there is significant potential to improve them. For example, only $9.29\%$ of all debris-flow-related disasters were predicted in China between 2010 and 2020 \cite{china}. In Japan from 2008 to 2013, only $3\%$ and $9\%$ of debris-flow-related disaster evacuation warnings were correct, based on two different models \cite{japan}. Failure to issue an evacuation warning prior to a debris flow can lead to multiple casualties, while frequent false alerts may affect the trustworthiness of evacuation orders.

This study demonstrates that a data-driven machine learning approach has the potential to improve the warning system in Taiwan. On the $85\%$ training set, we applied five machine learning models (Table \ref{table:cv_scores_ml_models}) and found the best performance from a random forest model (RF) using the hourly rainfall in the most recent 96 hours (or longer) as inputs. On the $15\%$ test set, the RF outperformed the ETM and the HM under all circumstances. The RF had the largest areas under the ROC (AUROC) and PR curves (AUPRC) (Fig. \ref{fig:roc_pr}), indicating better performance as a classifier. A data-driven machine learning approach can reduce false negatives and false positives in the debris flow occurrence predictions in Taiwan.

For the RF using the hourly rainfall in the most recent 96 hours as inputs, hourly rainfall in the most recent hour was the input with the heaviest weight in the debris flow prediction (Fig. \ref{fig:top12_shap_rf}). The RF using the hourly rainfall in the most recent hour had decent model performance (AUPRC=0.409), which demonstrates that the most recent hourly rainfall is an important indicator for debris flow occurrence within the upcoming 12 h (Supplementary Information). The 12 most important data inputs in the best-performing RF are mostly the hourly rainfall values in the most recent 14 hours. More recent hourly rainfall values are generally more important input data for debris flow prediction while the influence of past hours dropped off quickly. However, the trend shown in a SHAP plot (Fig. \ref{fig:top12_shap_rf}) only means the correlations between data inputs and predictions rather than the causalities \cite{molnar2022, lundberg2020local2global, lundberg2017unified} in the physical and hydrological mechanism of debris flow occurrence. Those correlations and the patterns of influence of data from different points in the past may be useful in refining the equation of the EAR (equation (\ref{eq:effective_antecedent_rainfall})).

Regardless of whether the EAR was added as input data, we found that the RF using hourly rainfall as inputs achieved similarly competitive model performance using simpler input data combinations, compared with other types of input data combinations (Supplementary Information). It is because the hourly rainfall values constitute 100\% of the available information and the RF works well as long as it has all of the information. Moreover, the hourly rainfall in the most recent 14 hours is so important that there is no way for a daily or a mix of both daily and hourly input data combination to do as well as hourly one (Fig. \ref{fig:top12_shap_rf}). The strong importance of the hourly rainfall from the most recent 14 hours in our analysis also suggests that the daily resolution in the EAR may be a large part of its limitation.

In addition, the more hourly rainfall values were used as inputs, the better model performance (Supplementary Information). However, the model performance hit a performance plateau when the hourly rainfall in most recent 96 hours were used as inputs. Adding more hours did not improve the predictions, mainly because hourly data from more than 96 hours in the past did not have much contribution to a debris flow occurrence. These hourly data from more than 96 hours ago can also be a source of noise for overfitting. The degradation of model performance from adding extra hourly data inputs was, however, small.

Among the RF considering the EAR as one of the inputs, the EAR was one of the most influential data inputs, which may validate the utility of EAR and reinforce the insight behind the traditional approaches. However, the AUPRC of the RF using the EAR as the only input data was much smaller than the other 27 input data combinations which consider daily, hourly, or both daily and hourly rainfall values in addition to the EAR (Supplementary Information). In other words, since the EAR lost the details in the daily and hourly aggregate of rainfall, using more information in addition to the EAR can significantly enhance the accuracy of predictions for debris flow occurrences. In contrast, adding the EAR as one of the inputs only slightly improved model performance, which validates that machine learning models can provide accurate predictions only using hourly rainfall values without relying on the traditional approaches such as the EAR from the ETM. The marginal improvement from adding the EAR is probably because the EAR is partially based on daily rainfall data. So, the temporal resolution in EAR is too coarse for good predictions. Another possible reason is that the uniform exponential weighting factor (0.7) in the EAR of the ETM is an incorrect choice. In future work, it is worthwhile trying different weighting factors to further investigate this issue.

The AUPRC values of the RF using the 55 input data combinations on the $85\%$ training set, including the 6 combinations in Table \ref{table:cv_scores_ml_models}, were close to each other, given that the variance of these AUPRC values was 0.004 (Supplementary Information). The low variance (0.004) demonstrates the strength of our machine learning models, which are insensitive to the details of the way the input data are organized and composited, regardless of whether the input data is EAR, daily, hourly, or both daily and hourly rainfall values. Most importantly, the average AUPRC of the RF using those 55 input data combinations on the $15\%$ test set is 0.257, which is about twice as large as the AUPRC of the ETM (0.141) and the HM (0.134). To get a more robust ``generalization'' evaluation, we used the whole dataset ($85\%$ training set $+$ $15\%$ test set) to do a 10-fold cross-validation on the RF using the hourly rainfall in the most recent 96 hours as inputs, and the mean of the AUPRC was 0.618. In contrast, the AUPRC of the ETM and the HM on the whole dataset were 0.207 and 0.298, respectively. This shows the robustness of the RF, given the fact that it performed well in examples not used in the training process.

In addition to pursuing optimal performance in debris flow occurrence prediction, this study provides decision-makers with a comprehensive overview of operating points and the multi-objective analysis of mitigating the undesired consequences of false alerts and warning omissions, which are conflicting goals in evacuation alerts. Since there is no clear pattern between tweaking the training weights of machine learning models and the impacts of minimizing the false alerts and warning omissions, the patterns in Fig. \ref{fig:multi_objectives} are not useful for evacuation planning. However, with sufficient data, the lines in Fig. \ref{fig:multi_objectives} would provide clearer patterns, and disaster managers can use the trade-offs between those conflicting goals as tools to decide an appropriate training weight for the future evacuation operations under a preferred level of objective, such as false omission rate. Disaster managers can also make the trade-offs between the number of debris flows predicted or missed by tweaking the probabilistic thresholds in the machine learning models (Fig. \ref{fig:Test_threshold_predict_analysis}). While a lower threshold allows disaster managers to issue more warnings and thus predict more debris flows, the precision is expected to be sacrificed while the threshold is decreasing.

One limitation of this study is that we could not include all actual negative rainfall events due to the following reasons. First, anomalies in weather stations caused interruptions in some data records. Second, the case reports of debris-flow-related disasters in Taiwan recorded the EAR threshold at the time of the debris flow. Third, the EAR threshold list published by the SWCB each year includes only two reference stations for each village, which does not cover all stations considered in our current study \cite{major}. Therefore, to compare performance of different models using the official EAR thresholds, we can include only the negative events that occurred at a station that had a debris flow in the same year. In Methods, we estimated how many datasets would be collected if all negative events were included, and future work should investigate the impact of making the prevalence of debris flow in the datasets used in training machine learning models match the prevalence in the real world.

There is potential to further improve and assess the performance of machine-learning-based debris flow warnings. First, it may be useful to add site-specific information on topography, soils, and site history, which may provide additional information useful for predicting debris flow events at different locations. Second, more sophisticated machine learning models for time series forecasting could be used. Third, it would be useful to explore the implications of issuing warnings with different lead times, integrating both the accuracy of the prediction and the consequences of receiving them further from or closer to the debris flow event. Given the promising results of our study, there is potential to adapt our data-driven machine learning models to other debris-flow-prone sites across the globe. Because global climate change is rapidly changing the precipitation pattern and intensity, it will be critical to train and test the machine learning models using the latest data or to potentially integrate modeled precipitation data into the analysis.

%% file: methods.tex
\section*{Methods}

\subsection*{Datasets}
We obtained the hourly rainfall data from the Central Weather Bureau (CWB) \cite{cwb} and the Data Bank for Atmospheric and Hydrologic Research \cite{databank}. Then, we compiled the rainfall data along with the debris flow data to form the positive and negative datasets using the following procedure.

\subsubsection*{Positive events}
We retrieved the dates and times of 104 debris flow events (positive events), which occurred in Taiwan from 2015 to 2020, from the case reports of debris-flow-related disasters published by the Soil and Water Conservation Bureau (SWCB) of Taiwan \cite{major}. Each case report was published as a PDF file, which we manually parsed to obtain the dates and times of the debris flow events. We then paired the hourly rainfall data with the debris flow data, where each hourly rainfall value was either assigned a negative label if no debris flow occurred within 12 hours or a positive label if a debris flow occurred within 12 hours. Each positive-event dataset consists of 7 days of hourly rainfall data prior to the main rainfall event during which a debris flow occurred, followed by the main rainfall event with a debris flow occurrence, and then appended with approximately 24-h hourly rainfall data subsequent to the main rainfall event, for a total of 210 hours of rainfall data on average. Finally, these 104 positive datasets were randomly split into 89 training sets and 15 test sets for our machine learning models.

\subsubsection*{Negative events}
To obtain data for negative events, when rainfall events did not cause debris flow, we considered only the weather stations that have been used as reference stations of rainfall records in the case reports of debris-flow-related disasters in Taiwan from 2015 to 2020 \cite{major}. We limited our negative events to those that came from the same year when a debris flow occurred because these were the only years for which the thresholds of EAR for the ETM were published in the case reports of debris-flow-related disasters. For example, a debris flow occurred at the Shangguguan Station on May 19, 2019 for which we compiled one positive dataset. Then, we collected all negative events that only contain rainfall events which lasted at least two consecutive hours but no debris flow occurred between January 1, 2019 and December 31, 2019. 

Originally, we obtained 2187 negative events in which rainfall lasted at least two consecutive hours but no debris flow occurred from 2015 to 2020. They were further consolidated into 533 unique negative datasets. Similar to the time series structure of a positive dataset, each negative dataset also has around 210 hours of rainfall data that consist of 7 days of antecedent hourly rainfall data, followed by a main rainfall event during which no debris flow occurred, and then appended with approximately 24-h hourly rainfall data after the main rainfall event. Moreover, there was no time overlap between the time series of a negative dataset and the time series of a positive one. The time windows of negative events did not overlap. If two of those 2187 negative events were right next to each other in a time series, they would end up on the same negative dataset, not two. By doing so, we consolidated all 2187 negative events into 533 negative datasets. Finally, these 533 negative datasets were randomly split into 456 training sets and 77 test sets for our machine learning models using the same ratio of training sets (85\%) and test sets (15\%) as for the positive datasets. 

Although there were many negative datasets in the real world, we included only a subset because of the limitations mentioned in Discussion. To better understand the number of negative datasets we did not include in this study, we estimated the number of all possible negative events in the real world as follows. First, we used 58 stations and six years of data or 348 station years. Then, the rainy season is about 5 months (or 20 weeks) long in Taiwan \cite{john2010diurnal}. Finally, 348 station years times 20 weeks per year of the rainy season is 6960 station weeks of rainy season data. However, we only had 69 station years in our datasets because we only included the negative events at the same station that had a debris flow in the same year. Therefore, 69 station-years times 20 weeks per year of the rainy season is 1380 station weeks of rainy season data. In reality, we have 637 datasets, each of them is an approximately week-long period, and 637 is close to the 1380 station weeks we should have. After considering the anomalies in rainfall data records and the annual changes in the locations of weather stations, we only collected 637 datasets (104 positive datasets and 533 negative datasets) rather than 1380.

\subsection*{Taiwan models}
\subsubsection*{The definition of the effective accumulated rainfall (EAR)}
In the existing rainfall-based debris flow warning model in Taiwan (the ETM), the effective accumulated rainfall (EAR) ($R_{eff}(t)$ measured in mm at time (hour) $t$ is calculated through equation (\ref{eq:effective_antecedent_rainfall}). It is the sum of two terms.
\begin{itemize}
    \item The first term, $R_{main}(t)$, is the accumulated rainfall at time (hour) $t$ of the main rainfall event that may contribute to a debris flow \cite{jan2018txt}. A main rainfall event is defined as follows. It starts when hourly rainfall exceeds 4 mm and terminates at the hour when the hourly rainfall is still above 4 mm, and then it drops below 4 mm in at least the next six hours \cite{jan2018txt}.   
    \item The second term is the antecedent precipitation index that expresses how wet the watershed is when the main rainfall event begins. The higher the antecedent precipitation index, the higher the potential that rainfall will result in surface runoff and debris flow.  $R_i$ is the daily rainfall on the $i-th$ day before the first hour of the main rainfall event and $\alpha^{i}$ is a weighting factor that signifies that the effect of preceding rainfall decays with passing time \cite{jan2018txt}. The ETM uses $\alpha = 0.7$ \cite{jan2018txt}. 
\end{itemize}

We calculate the EAR until the hourly rainfall dropped below 4 mm and lasted over 6 h. 

\begin{equation} \label{eq:effective_antecedent_rainfall} {R_{eff}(t) = R_{main}(t)+ \sum_{i=1}^{7} \alpha^{i} R_i} \\
\end{equation}

\subsubsection*{The existing Taiwan model (ETM)}
When the EAR threshold is exceeded at time (hour) $t$ in the existing rainfall-based debris flow warning model in Taiwan (the ETM), a mandatory evacuation order is issued \cite{jan2018txt}. Based on the frequency of debris flow occurrences in recent years, hydrologists and disaster managers adjust the EAR threshold, which can range from 200 mm to 600 mm in increments of 50 mm \cite{jan2018txt}, for each debris-flow-prone river catchment every year. A smaller EAR threshold value implies that less precipitation is required to trigger a debris-flow evacuation warning \cite{jan2018txt}.

\subsubsection*{The homogeneous model (HM)}
The difference from the ETM is that the HM applies the same threshold value of EAR to all catchments.

\subsection*{Machine learning models}
We framed the machine learning task as a binary classification problem, where the goal is to predict whether a debris flow will occur within the upcoming 12 hours given a sequence of historical rainfall data. The input variables were a combination of EAR, daily rainfall values, and hourly rainfall values. Each sequence is paired with a label indicating whether a debris flow will occur within the upcoming 12 hours. To make a prediction on an hourly basis, we shift the time window of the sequence by 1 hour and recomputing the input variables and corresponding labels. In total we included 29,304 pairs of rainfall sequences and labels in the dataset that was used to train and evaluate the machine learning models. Each machine learning model has a different functional form and parameters that are optimized to minimize error, where the error criterion and optimization procedure also differ from model to model. We split the dataset into 25,077 pairs as the 85\% training set to learn the machine learning model parameters and 4,227 as the 15\% test set to evaluate the best machine learning model on previously unseen data.

\subsubsection*{Input features}
We used hourly rainfall values as the input feature in all machine learning models. In the Supplementary Information, we also considered different input data combinations of EAR, daily, hourly, or the mixed setting of both daily and hourly rainfall as an ablation study to measure the impacts of different input data on the performance of the random forest model (RF), which was the best performing machine learning model in this study. Similar to equation (\ref{eq:effective_antecedent_rainfall}), when we formed the daily inputs, we used the daily sum of rainfall on the $i-th$ day from 00:00 to 24:00 with or without the weighting factor $0.7^{i}$ prior to the time of prediction. However, no matter whether the weighting factor of $0.7^{i}$ was applied to the daily sums of rainfall in the RF or not, the importance ranking of each input feature was exactly the same and the model performance (AUPRC) was almost the same. This is expected because tree-based machine learning models, such as the RF, are intrinsically not affected by the absolute value of the input features, only by the relative differences in input feature values between different examples. So, even when the weighting factor of $0.7^{i}$ was applied, the relative differences in input feature values between different examples were the same to the RF.

As for the hourly inputs, we used hourly rainfall values, which could range from 1 h to 168 h prior to the time of prediction (Supplementary Information). For the mixed setting, we used both daily and hourly rainfall values as input but they did not overlap in time. We used these three different settings of input data to train five machine learning models, each of which makes a prediction (positive or negative) every hour.

We compiled the hourly precipitation data along with labeling the hours of having a debris flow occurrence within $L$ h as $positive$ and otherwise as $negative$ in each rainfall event. Specifically, if we wanted to make a prediction with lead time = $L$ h, we labeled the hour when the debris flow occurs as positive and added additional positive labels for $L$ h before the debris flow happened. In other words, if our machine learning models predicted $positive$ any time within the lead time $L$ h before the debris flow occurred, the prediction counted as correct, which was the same way we used to count the number of correct predictions in the ETM and the HM. In this study, we set up 12 h as the lead time.

We paired input features with a ground truth output label for each hour to train our machine learning models to \textit{learn} the relationship between rainfall values (referred to as \textit{features} \cite{ng2000cs229}) and whether the debris flow occurred within the lead time $L$ h (referred to as \textit{targets} or \textit{labels} \cite{ng2000cs229}).

\subsubsection*{Models}
Here we review the high-level definitions and basic concepts for the machine learning classification algorithms in this study.

\paragraph*{Logistic Regression}
We used a logistic regression (also known as softmax regression) model for binary classification to  predict the probability of whether a debris flow would occur within the upcoming 12 h. Logistic Regression uses a linear function with learned parameters that produces a continuous score. Then, this continuous score is fed into a softmax function to obtain a value between 0 and 1 (in the binary setting, the softmax function is referred to as a logistic function, or a sigmoid function) \cite{kleinbaum2002logistic, pampel2020logistic}. The parameters of the function were learned using the LBFGS optimization algorithm \cite{morales2011remark, scikit-learn, zhu1997algorithm, byrd1995limited}. During cross-validation, we hyperparameter tuned the regularization penalty (no regularization or L2 regularization) and corresponding coefficient (0.001, 0.01, 0.1, 1, 10, 100, 1000). 

\paragraph*{Multi-layer Perceptron (MLP)}
A Multi-layer Perceptron (MLP) (a type of Artificial Neural Network) is a composition of linear functions separated by nonlinear functions, where the linear functions have learnable parameters that are optimized on the training data \cite{taud2018multilayer, bishop1995neural}. The series of functions in the network are referred to as layers, where the layers besides the first and the last are referred to as hidden layers, and the elements of the outputs of each layer are referred to as neurons. Similar to a logistic regression model, an MLP operates on the input to produce a continuous score and then the score was fed into a logistic function to predict the probability of whether a debris flow would occur within the upcoming 12 h. The parameters were learned using the Adam optimization algorithm. During cross-validation, we hyperparameter tuned the number of hidden layers (1, 2) and their sizes (5 to 30 neurons), the nonlinear functions (Tanh, ReLU), and the learning rate for Adam (0.01, 0.001, 0.0001).

\paragraph*{Random forest}
Random forests are ensemble algorithms consisting of several randomized decision trees, which have been widely applied to regression and classification problems \cite{biau2016random, cutler2012random, biau2012analysis, breiman2001random}. Decision trees learn to threshold the input variables in order to partition the training dataset into different groups, where the labels of the groups are used to make new predictions. Random forests build an ensemble of independent decision trees whose outputs are combined using an average that has been shown to enhance accuracy in prediction and prevent overfitting \cite{scikit-learn, hastie2009random}. During cross-validation, we hyperparameter tuned the number of decision trees (10, 40, 70, 100), the maximum depth of the decision trees (no max depth, 1, 2, 6, 15, 39, 100), and the minimum number of samples per group (1, 2, 4).

\paragraph*{Extreme Gradient Boosting (XGBoost)}
Extreme Gradient Boosting (also referred to as XGBoost) is also an ensemble algorithm consisting of several decision trees, where the trees are learned sequentially using gradient boosting and scalably using several algorithmic advancements \cite{chen2016xgboost, chen2015xgboost, friedman2001greedy, friedman2000additive}. XGBoost has been widely used on a wide variety of tasks and has led to successful solutions in many machine learning competitions. During cross-validation, we hyperparameter tuned the number of decision trees (10, 40, 70, 100), the maximum depth of the decision trees (no max depth, 1, 2, 6, 15, 39, 100), the minimum weight per child (1, 2, 4), and the learning rate (0.001, 0.01, 0.1, 1.0).

\paragraph*{Light Gradient Boosting Machine (LightGBM)}
Light Gradient Boosting Machine (LightGBM) is also a gradient boosting-based ensemble of decision trees, but retains similar accuracy with higher training efficiency compared to a typical Gradient Boosting Decision Tree (GBDT) by focusing on a small portion of data instances with larger gradients and bundling mutually exclusive features to reduce the number of features required to achieve similar accuracy \cite{ke2017lightgbm}. During cross-validation, we hyperparameter tuned the number of decision trees (10, 40, 70, 100), the maximum depth of the decision trees (no max depth, 1, 2, 6, 15, 39, 100), the minimum weight per child (1, 2, 4), and the learning rate (0.001, 0.01, 0.1, 1.0).

\subsection*{Evaluation}
\subsubsection*{Point metrics}
\begin{figure}[H]
\includegraphics[width=0.5\textwidth]{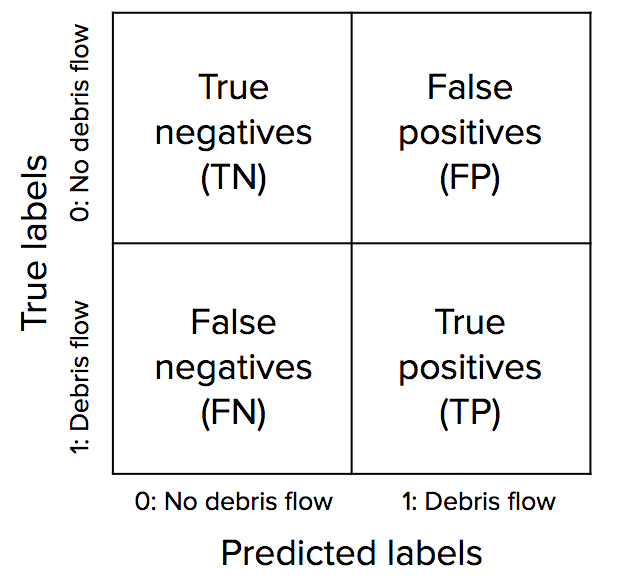}\centering
\caption{\textbf{Confusion matrix of this study.}}\label{fig:confusion_matrix}
\centering
\end{figure}

We used a confusion matrix to evaluate the performance of a classification model \cite{tharwat2020classification}. This matrix (Fig. \ref{fig:confusion_matrix}) comprises the number of true positives (TP), true negatives (TN), false positives (FP), and false negatives (FN). We define the four possible outcomes as follows: A true positive (TP) event means that the model predict $positive$, meaning a debris flow is predicted to happen within the upcoming 12 h, and indeed, a debris flow occurs within 12 h. In contrast, if the model prediction is  $negative$, meaning a debris flow would not happen within the upcoming 12 h, and no debris flow occurs within 12 h, we refer to this as a true negative (TN) event. If the model prediction is $positive$ but the true label is $negative$, it is a false alert or a false positive (FP) event. Similarly, if the model prediction is $negative$ but the true label is $positive$, the model fails to issue an early warning within 12 h before the debris flow occurs, and we refer to this as a false negative (FN) event.

\subsubsection*{Summary metrics}
To quantify the overall performance of our machine learning models and the ETM using various classification thresholds, we used Receiver Operating Characteristic curves (ROC curves; Fig. \ref{fig:roc_pr}a) and Precision-Recall curves (PR curves; Fig. \ref{fig:roc_pr}b). To comprehensively visualize the impacts of various critical accumulated rainfall values in the ROC curves and the PR curves, we decreased the predetermined threshold values of EAR until 0, where the ETM and the HM always issue evacuation alerts, and increased the values until the maximum computed value until the models did not issue any evacuation alerts. Similar to what we did in the ETM and the HM, we constructed the curves by varying $P_{predict}$ from 0 to 1. At each point along the curve, we evaluated the false positive rate (FPR, equation (\ref{eq:FPR})), true positive rate (TPR; also referred to as $recall$ and calculated as equation (\ref{eq:TPR})), and precision (also referred to as $positive\ predictive\ value\ (PPV)$, equation (\ref{eq:precision})). The ROC curve plots FPR against TPR and does not depend on the prevalence of positive outcomes. The PR curve plots precision against recall and is affected by the prevalence of positive outcomes.

\begin{equation}\label{eq:FPR}
\begin{split} 
False\ Positive\ Rate\ (FPR)\ 
&=\ \frac{False\ Positives\ (FP)}{False\ Positives\ (FP)\ + True\ Negatives\ (TN)}\\ 
&=\ \frac{Number\ of\ False\ Alerts}{Total\ Number\ of\ No\ Debris\ Flow\ Occurrences\ within\ 12\ h}\\
\end{split}
\end{equation}

\begin{equation}\label{eq:TPR}
\begin{split} 
True\ Positive\ Rate\ (TPR)\
&=\ Recall\\
&=\ \frac{True\ Positives\ (TP)}{True\ Positives\ (TP)\ + False\ Negatives\ (FN)}\\ 
&=\ \frac{Number\ of\ Correct\ Predictions\ for\ Debris\ Flow\ Occurrences\ within\ 12\ h}{Total\ Number\ of\ Debris\ Flow\ Occurrences\ within\ 12\ h} 
\end{split}
\end{equation}

\begin{equation}\label{eq:precision}
\begin{split} 
Positive\ Predictive\ Value\ (PPV)\
&=\ Precision\\
&=\ \frac{True\ Positives\ (TP)}{True\ Positives\ (TP)\ + False\ Positives\ (FP)}\\ 
&=\ \frac{Number\ of\ Correct\ Positive\ Predictions}{Number\ of\ Positive\ Predictions\ for\ Debris\ Flow\ Occurrences\ within\ 12\ h} 
\end{split}
\end{equation}

\begin{equation}\label{eq:specificity}
\begin{split} 
Specificity\
&=\ True\ Negative\ Rate\ (TNR)\\
&=\ \frac{True\ Negatives\ (TN)}{True\ Negatives\ (TN)\ + False\ Positives\ (FP)}\\ 
&=\ \frac{Number\ of\ Correct\ Predictions\ for\ No\ Debris\ Flow\ Occurrences\ within\ 12\ h}{Total\ Number\ of\ No\ Debris\ Flow\ Occurrences\ within\ 12\ h} 
\end{split}
\end{equation}

\begin{equation}\label{eq:FNR}
\begin{split} 
False\ Negative\ Rate\ (FNR)\ 
&=\ \frac{False\ Negatives\ (FN)}{False\ Negatives\ (FN)\ + True\ Positives\ (TP)}\\ 
&=\ \frac{Number\ of\ Missing\ Warning\ Incidents}{Total\ Number\ of\ Debris\ Flow\ Occurrences\ within\ 12\ h}\\
&=\ 1 - True\ Positive\ Rate\ (equation\ (\ref{eq:TPR}))
\end{split}
\end{equation}

\begin{equation}\label{eq:FDR}
\begin{split} 
False\ Discovery\ Rate\ (FDR)\ 
&=\ \frac{False\ Positives\ (FP)}{False\ Positives\ (FP)\ + True\ Positives\ (TP)}\\ 
&=\ \frac{Number\ of\ False\ Alerts}{Number\ of\ Positive\ Predictions\ for\ Debris\ Flow\ Occurrences\ within\ 12\ h} 
\end{split}
\end{equation}

\begin{equation}\label{eq:FOR}
\begin{split} 
False\ Omission\ Rate\ (FOR)\ 
&=\ \frac{False\ Negatives\ (FN)}{False\ Negatives\ (FN)\ + True\ Negatives\ (TN)}\\ 
&=\ \frac{Number\ of\ Missing\ Warning\ Incidents}{Number\ of\ Predictions\ of\ No\ Debris\ Flow\ Occurrences\ within\ 12\ h} 
\end{split}
\end{equation}

The number of true negatives can only decrease when the probabilistic threshold decreases, and the model thus issues warnings more often because the probabilistic threshold is exceeded more easily. Therefore, the false-positive rate monotonically increased with the true positive rate in the ROC curves (Fig. \ref{fig:roc_pr}a). The jaggedness in the ROC curves is in part due to certain thresholds leading to sharp changes in metrics, and in part due to the number of test sets being relatively small. As for the PR curves, the zigzag shape of non-monotonicity was expected because we used precision to evaluate the prediction performance, which is explained as follows (Fig. \ref{fig:roc_pr}b). The denominator of precision is the number of positive predictions, and the numerator is the number of true positives (equation (\ref{eq:precision})). In other words, if a probabilistic threshold decreased and the machine learning model predicted many true positives and kept the number of false positives almost the same, both the numerator and the denominator of precision increased. So, the precision climbed up to a higher level slowly, while recall was increasing (Fig. \ref{fig:roc_pr}b). When recall was around 0.34 and 0.49 in the ETM and the HM, respectively, the precision drastically dropped because any thresholds beyond this tipping point started to generate more false positives than true positives while recall either stayed the same or just slightly increased due to the mild increase in the number of true positives.